# Durghotona GPT: A Web Scraping and Large Language Model Based Framework to Generate Road Accident Dataset Automatically in Bangladesh

MD Thamed Bin Zaman Chowdhury
*Department of Civil Engineering*
*Bangladesh University of Engineering and Technology*
Dhaka, Bangladesh
zamanthamed@gmail.com

Moazzem Hossain
*Department of Civil Engineering*
*Bangladesh University of Engineering and Technology*
Dhaka, Bangladesh
moazzem@ce.buet.ac.bd

Md. Ridwanul Islam
*Department of Civil Engineering*
*Bangladesh University of Engineering and Technology*
Dhaka, Bangladesh
mdridwan1275@gmail.com

*Abstract*—Road accidents pose significant concerns globally. It leads to large financial losses, injuries, disabilities and societal challenges. Accurate and timely accident data is essential for predicting and mitigating these events. This paper presents a novel framework named 'Durghotona GPT' that integrates web scraping and Large Language Models (LLMs) to automate the generation of comprehensive accident datasets from prominent national dailies in Bangladesh. The authors collected accident reports from three major newspapers— Prothom Alo, Dhaka Tribune and The Daily Star. The collected news was then processed using the newest available LLMs. These LLMs are: GPT-4, GPT-3.5 and Llama-3. The framework efficiently extracts relevant information, categorizes reports and compiles detailed datasets. Thus, this framework overcomes limitations of manual data collection methods such as delays, errors and communication gaps. The authors' evaluation demonstrates that Llama-3, an open-source model, performs comparably to GPT-4. It achieved 89% accuracy in the authors' evaluation. So, it can be considered as a cost-effective alternative for similar tasks. The results suggest that the framework developed by the authors can drastically enhance the quality and availability of accident data. As a result, it can support critical applications in traffic safety analysis, urban planning and public health. The authors also developed an interface for 'Durghotona GPT' for easy use as a part of this paper. Future work will focus on expanding data collection methods and refining LLMs to further increase dataset accuracy and applicability.

I. INTRODUCTION

Road accidents represent a significant concern worldwide. It impacts the economy, public health and social stability. In Bangladesh, the frequency and severity of road accidents are alarming. So, it is necessary to collect data efficiently and analyze for effective intervention and prevention. Accurate and timely data on road accidents is crucial for predicting trends, identifying high-risk areas and implementing measures to enhance road safety. Traditionally, accident data collection has relied heavily on manual methods. These methods involve police reports, hospital records and accident research organizations. However, these methods are often plagued by delays, inaccuracies and incomplete data. These problems limit their utility in real-time analysis and decision-making.

With the emergence of machine learning (ML) as well as artificial intelligence (AI), there has been a surge in research focused on using these technologies for accident analysis. Studies have explored various ML models to predict accident severity and analyze contributing factors using historical data from multiple sources. While these efforts have yielded valuable insights, they often lack the most recent data. Having latest data is vital for accurate prediction and continuous monitoring of changing accident patterns. Moreover, the manual extraction and organization of data introduce several challenges, including errors, delays, additional costs and communication gaps between researchers and data providers.

To tackle these challenges, the authors in this paper presents a novel framework that integrates web scraping and Large Language Models (LLMs) to automate the generation of comprehensive accident datasets from prominent national newspapers in Bangladesh. The framework is named 'Durghotona GPT'. Newspapers can sometimes act as a good supplement to police report because of long waiting period to get the police reports in hand after gaining several clearances. Moreover, as shown in [1], police report can have significant missing data as large as 75 percent in certain scenarios. By utilizing web scraping techniques, one can very effectively collect the latest accident reports from online sources. This ensures a continuous flow of up-to-date information. The collected reports are then processed using modern available LLMs which include GPT-4, GPT-3.5 and Llama-3. These LLMs extract relevant details and compile detailed datasets. 'Durghotona GPT' not only increases the efficiency and accuracy of collecting data but also ensures scalability and adaptability to different sources and formats.

This paper's framework addresses the limitations of manual data collection methods and thus, provides a robust and scalable solution that significantly enhances the availability and quality of accident datasets. This, in turn, supports a wide range of applications in traffic safety analysis, urban planning and public health. By automating the extraction of accident data from newspapers, this framework can facilitate more accurate and timely analyses.

This ultimately contributes to improved road safety and reduced accident-related impacts.

The remaining sections of the paper is organized following the sections mentioned: Section II discusses related works done in the fields of web scraping and LLMs for data extraction. Section III describes the architecture and implementation of proposed framework. Section IV discusses the methodology. After that Section V highlights results which demonstrates the efficacy of this approach. At the end, Section VI draws necessary conclusion of the paper with discussions on potential applications and future directions for research.

## II. RELATED WORKS

The authors could not find any significant works on automating accident dataset collection in context of Bangladesh. Recent studies have focused on predicting road accidents and analyzing factors that contribute to road accident using ML (machine learning) models. The authors in [2] conducted an analysis of road crashes in Bangladesh which uses Decision Tree, Naïve Bayes, AdaBoost and K-Nearest Neighbors (KNN) models to classify accident severity. Their dataset, limited to accidents up to 2015. So, they lacked recent data necessary for understanding current trends and factors influencing road accidents. The authors in [3] also addressed road accident prediction in Bangladesh, utilizing data from the National Traffic Accident Report 2007 and the Bangladesh Road Transport Authority (BRTA). This study highlighted the challenges of accessing comprehensive and up-to-date datasets, which is crucial for accurate prediction and analysis.

The closest work the authors could find similar to this study is in [4] where the authors in [4] extracted accident information from Chinese media websites utilizing Google's BERT (Bidirectional Encoder Representations from Transformers). However, the performance of modern LLMs, for example, GPT-3.5, GPT-4 and Llama-3 in this task is not explored and the prospective of using latest web scraping tools is not discussed in [4]. Moreover, the study in [4] was done in context of China whereas in context of Bangladesh, this is a completely unexplored field.

Other works also use either web scraping or LLMs to collect or extract information but a combination of these two is not explored in them. In [5], the authors again used Google's BERT LLM for identification of actual WWD (wrong-way driving) accidents from police narrative reports. In [6], authors utilized web scraping techniques to automatically collect and aggregate factual claims from articles sourced from online newspaper. Similarly, in [7], authors explored the prospective of analyzing the data following Robot Process Automation (RPA) technique using Web Scraping and UI Path Studio (a Windows-based desktop tool that enables the creation of cross-platform automations and provides advanced testing and debugging capabilities.) which explores automated data collection of COVID-19 cases using web scraping. However, application of web scraping in accident research is not explored.

## III. PROPOSED FRAMEWORK

Fig .1 illustrates the ultimate objective of this paper.

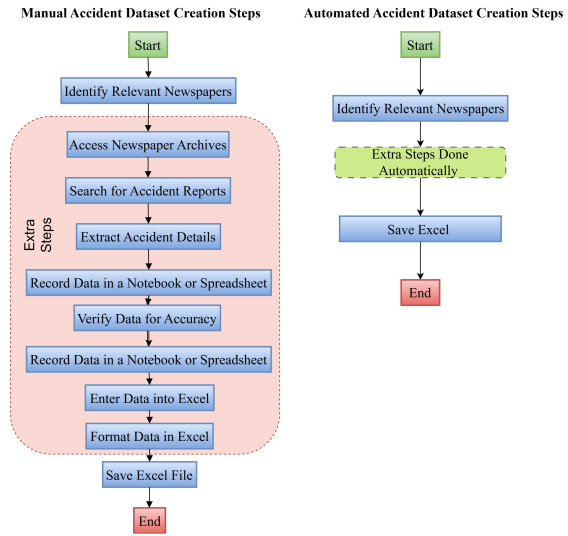

Fig.1. Manual vs Automated method of accident dataset generation.

In this paper, the authors are proposing a framework that can completely remove all the inconveniences of creating accident dataset manually. The framework consists of two primary components: A. Data Collection using 'Web Scraping' and B. Data Processing using 'Large Language Models (LLM)'. Two separate Python programs work together to achieve this. The following subsections provide a simple overview of the overall working process.

### A. Web Scraping:

Web scraping refers to a process that can automatically extract data from websites. It involves utilizing software tools known as web scrapers to access web pages, parse the HyperText Markup Language (HTML) content and extract useful information. Web scraping can be employed to serve different purposes, such as gathering data for market research, monitoring prices, aggregating information. In this case, accident news from multiple sources were collected using web scraping. Fig. 2 illustrates the overall working process of web scraping in this framework. It is imperative to make sure that the ethical and legal aspects of web scraping including adhering to a website's 'robots.txt' file and terms of service is followed. 'robots.txt' refers to a text file used by websites to instruct web crawlers and other automated agents. Typically, these automated agents are search engines. It is part of a protocol named Robots Exclusion Protocol (REP), a standard used by websites to inform web crawlers which pages should not be processed or analyzed. Since the newspaper websites being scraped in this project are renowned dailies providing free and opensource information, this project is not bound legally.

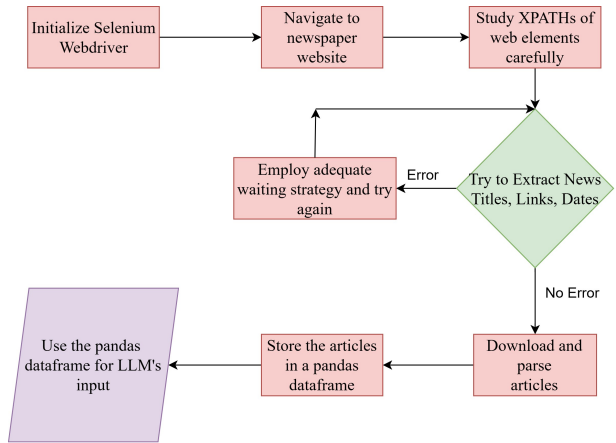

Fig. 2. The working mechanism of web scraping

*B. Data Processing Using LLMs:*

LLM stands for Large Language Model. LLMs are artificially intelligent models that can not only understand but also generate texts that are very similar to humans. A large size of text data is used to train these models and they can perform text analyzing tasks. Some examples of these jobs are- completing texts, answering user's questions and language translation. The workflow of data processing is given in Fig. 3 below:

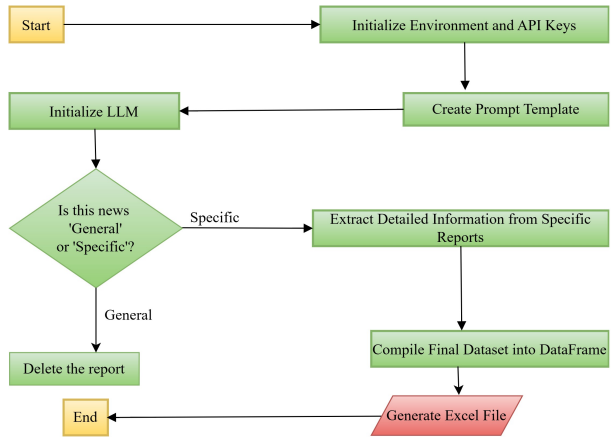

Fig. 3. Working mechanism of LLM data processing

## IV. METHODOLOGY

There are three parts in the methodology: A. Data Collection, B. Data Processing C. Validation of the framework. Each of the proposed methodology is discussed in detail in this section.

*A. Methodology of Data Collection using web scraping:*

Before even starting the process of web scraping, one must check if the selected website allows web scraping in the first place to avoid legal difficulties by inspecting the 'robots.txt' file associated to the website carefully. So, the authors first checked all the news websites' 'robots.txt' file. No websites had any instruction that restricts collecting news using web scraping. So, it is perfectly fine to go ahead with web scraping on these websites.

The authors used 'Selenium WebDriver', a popular python module for web scraping. The authors in [8] provide a detailed insight on the architecture of 'Selenium' framework. 'Selenium WebDriver' provides a mean of controlling an automated testing chrome browser using python scripts. First, the module is initialized. Then the python code navigates to the websites of the desired newspapers. Since different newspaper websites have different structures, a study of the websites' structure along with the XPaths of web elements is necessary. XPath used in Selenium can be referred to as Extensible Markup Language (XML) path used for the purpose of navigating HTML based pages. It is a language or syntax for utilizing an XML path expression to locate any element on a webpage. Xpath is a popular feature in many web scraping applications. For example, authors in [9] used Xpath for web scraping.

Then the code will attempt to extract the news titles, news links and publish dates of the news utilizing their respective XPaths similar to the approach in [10]. However, the program might run into various errors such as- No Such Element Exception. This type of errors can be easily bypassed by waiting for some times like 5 seconds to 10 seconds and then trying again. Once the program is successful in extracting news link, then as shown in [11], it downloads the news articles from each link using 'newspaper3k', another python module to extract limited information from newspaper. Lastly, the downloaded news article is saved in a pandas dataframe and returned for the LLM to use. In this manner, news from three renowned national dailies- "Prothom Alo", "The Daily Star" and "Dhaka Tribune" are extracted. Thus, this program can effectively collect latest accident news from newspaper without any intensive labor. The resulting dataframe will consist of three columns- 'News Title', 'News Link' and 'Publish Date'. With creation of this dataframe, the program has successfully collected accident related news along with their links which will be used in the next step.

*B. Methodology of Data Processing using LLM:*

For the second program, the authors followed a similar approach as shown in [12]. The second program first initializes all necessary environment and API (Application Programming Interface) keys. This program requires at least two API keys- first one is the OpenAI API and second one is the Groq API key. The author used OpenAI's API key for GPT-4 and GPT-3.5. For Llama3, Groq's API is used. The authors used LangChain to efficiently handle input and outputs in the LLM. LangChain is a popular platform for AI developers which supports a wide range of AI tools with an active developer community. OpenAI is the developing company of GPT (Generative Pre-Trained Transformers) based LLM models. On the other hand, Groq is a platform that has excellent software as well as hardware support. It also provides great quality, compute speed and energy efficiency. In this work, the authors used the base models without any finetuning to compare the results of the LLMs. The base setting of each LLM is shown in Table I.

TABLE I. BASE LLM SETTINGS

| Model Property | GPT-3.5 | GPT-4 | Llama-3 |
|---|---|---|---|
| Specific Name of the Model | GPT-3.5-turbo | GPT-4o | Llama-3-70b-8192 |
| temperature | 0.7 | 0.7 | 0.7 |
| max_retries | 2 | 2 | 2 |
| n | 1 | 1 | 1 |

There are various versions of these LLMs available suitable for different tasks. "Specific Name of the Model" refers to the exact version of the model used in this task. It should be noted that exact parameter count of GPT models is not disclosed by OpenAI officially yet. Llama-3 has a parameter count of 70 billion and context window of 8192 tokens. Temperature is a parameter that controls the randomness or creativity of an LLM. Max_retry refers to the maximum number of calls API will make in case of failure. Lastly, n value refers to number of responses generated by the LLM. In order to provide a fair ground for comparison, authors kept all the settings same (base setting) for all LLMs. Doing so, the accuracy of the LLMs can be compared effectively. It should be noted that inaccuracy of an LLM mostly happens due to "hallucination". Hallucination is an event where an LLM confidently generates wrong response. So, it is mandatory to measure an LLM's accuracy in a specific task.

Then, the program creates a chat prompt template between the system (LLM) and the user. Next, all the elements are combined to create a LangChain LLM chain which can be conveniently used to provide input and parse output. This 1st LLM chain contains a prompt that can categorize news reports into two types- 'General' and 'Specific'. Note that some reports are not about individual accidents but about some overall cases. For example, reports like- 'total number of accidents in April only', 'who is responsible for road accidents?', 'expert opinion on road accidents' etc. news are categorized as 'General' reports and these reports are then excluded from the dataframe. Now each news report is passed through this LLM chain one by one obtained from the previous program and the LLM is asked to categorize the reports either as 'Specific' or 'General' and then the 'General' reports are excluded from the dataframe. Now the dataframe is left with news reports only about 'Specific' accident incidents.

Next another 2nd LLM chain is created with necessary prompts to extract information from these 'Specific' news reports. These 'Specific' accident reports are also passed through the LLM chain one by one again and the following information are extracted- Accident date, Time, Number of injured, Number of killed, Location, Road characteristics, was there any pedestrian involved in the accident? What type of vehicles were involved in the accident? The response from the LLM for each report is then saved in a new Pandas dataframe and finally, the dataframe is exported as an excel file.

*C. Methodology of validating the framework:*

To measure the performance and accuracy of the LLMs and validate the framework, authors first generated a number of datasets using this framework. Then a number of news reports were inspected manually very carefully which served as a gold standard. Then this standard is compared against the framework's automatically generated dataset to assess its performance similar to the approach in [4]. The result of this analysis is shown in the next section.

The authors included all necessary source codes along with an interactive Graphical User Interface (GUI) for general purpose use of this project in [13] for anyone interested. It is completely free to use and can be reused following the terms and condition.

V. RESULTS AND DISCUSSION

The authors inspected a total of 195 news reports manually against three LLM's automatically generated datasets. Among the 195 reports, 60 reports were from "Prothom Alo", 64 reports were from "The Daily Star" and 71 reports were from "Dhaka Tribune". Fig. 3 shows LLM's accuracy in 'Prothom Alo', Fig. 4 shows accuracy in 'Daily Star' and lastly fig. 5 shows the accuracy in 'Dhaka Tribune'.

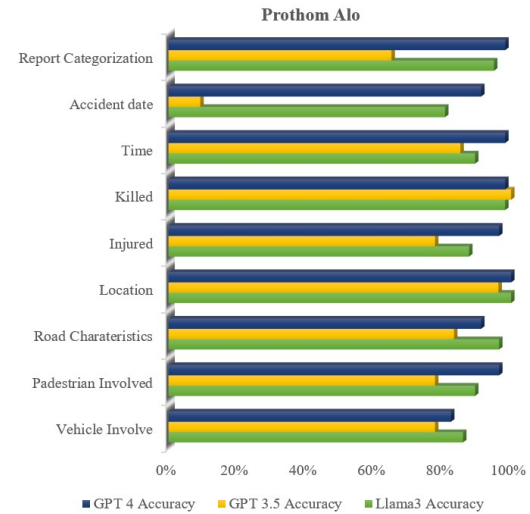

Fig. 3. Performance of LLMs on Prothom Alo

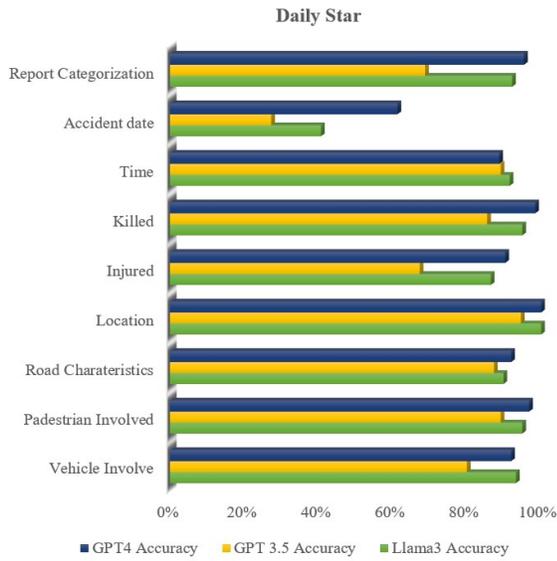

Fig. 4. Performance of LLMs on The Daily Star

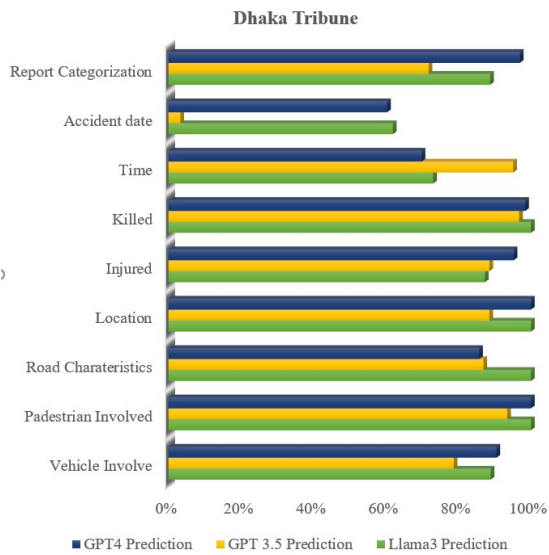

Fig. 5. Performance of the LLMs on Dhaka Tribune

The overall performances of the LLMs across all newspapers and across all parameters of the datasets is shown in Table II.

TABLE II. OVERALL LLM PERFORMANCES

| Llama 3 | | GPT-3.5 | | GPT-4 | |
|---|---|---|---|---|---|
| Correct | Wrong | Correct | Wrong | Correct | Wrong |
| 1450 | 177 | 1224 | 384 | 1499 | 145 |

Which LLM performed the best in this task can be found using Table I. The result is shown in Fig. 6.

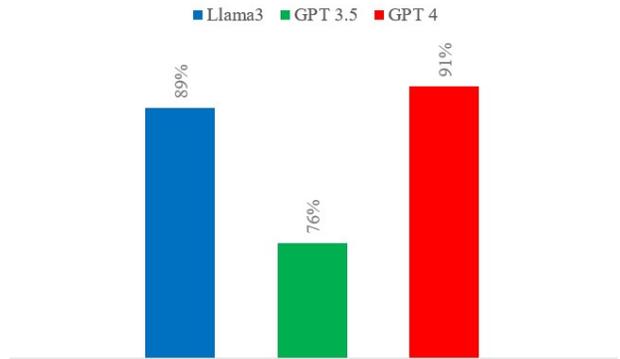

Fig. 6. Overall accuracy of the LLMs

It can be seen that GPT-4 performed the best with an impressive accuracy of 91%.. Llama-3 came in second with very close 89% accuracy and GPT-3.5 perfomed the worst with 76% accuracy. In [4], the authors used Google's BERT LLM model which yielded accuracy of over 80% in variables- 'Location', 'Roadway Type' and 'Deaths'. Moreover, it performed worse in some variables such as- 'Date' (64.1% accuracy), 'Time' (69.2% accuracy), 'Vehicles/Persons Involved' (51.2% accuracy) and 'Non-Fatal Injuries' (67.2%). However, in this paper's case, GPT-4 and Llama-3 performed better in these variables than Google's BERT except for accident 'Date'. None of the LLMs among BERT, Llama-3, GPT-4 and GPT-3.5 could not perform satisfactorily in 'Date' variable. In most of the other variables, GPT-4 and Llama-3 had an accuracy of over 90% which is a significant improvement over Google's BERT.

The authors chose these 3 LLMs specifically because although GPT-4 from OpenAI has state of the art performance, it is not free and quite costly. On the other hand, GPT-3.5 is used extensively and less costly than GPT-4. However, Llama-3 is the completely free opensource LLM from Meta. So, the authors tried to choose the LLMs based on cost and availability perspective. From the obtained results, it can be concluded that Llama-3 can be a very good option for these tasks as it is free and on par with GPT-4 in this task. Also, the obtained accuracy of both Llama-3 and GPT-4 is quite satisfactory and these LLMs can be effevtively used for automated dataset generation tasks or summarizing narratives in general.

VI. CONCLUSION

The swift evolution of Large Language Models (LLMs) and Artificial Intelligence (AI) in general has revolutionized the field of data analysis. It provides powerful tools for extracting and understanding information from vast textual datasets. In this study, the authors leveraged these advancements to efficiently and accurately collect road accident data. By comparing the performance of the three LLMs used in this paper, it was found that Llama-3 performs comparably to GPT-4 which is often considered state of the art. Although Llama-3 is a free open-source model, it achieved an impressive 89% accuracy. This paper also showed that this result is significant improvement over Google's BERT which is a frequently used LLM for data extraction from crash narratives. This makes Llama-3 a cost-effective alternative for similar tasks. Although it may have a low accuracy in one parameter- 'Accident Date', with additional prompt engineering and RAG (Retrieval Augmented Generation), it can be drastically improved. The automated framework developed in this study significantly enhances the collection and processing of accident-related news articles. It can provide a reliable and comprehensive dataset. Such data is crucial for conducting detailed analyses, informing policy decisions and implementing effective safety measures. Future work can explore expanding the scope of data sources and further improving LLM capabilities utilizing techniques such as- prompt engineering, agent-based systems, Retrieval Augmented Generation (RAG), LLM finetuning etc. Also, further work can be done on speeding up of the web scraping process. Additionally, as shown in [14], integrating this framework with a machine learning framework to build a complete pipeline can also be a further research topic for accident analysis. These researches will enrich the datasets and improve accuracy of the LLMs. It will contribute to advancements in traffic safety, urban planning and public health. This research underscores the transformative impact of AI in addressing practical challenges. Moreover, it highlights AI's role in improving the safety of public as well as quality of life. narratives in general.